\documentclass{Interspeech}

\interspeechcameraready

\usepackage{url}
\usepackage{multirow}
\usepackage{colortbl}
\usepackage{xspace}
\usepackage{hyphenat}

\newcommand{\customParagraph}[1]{{\noindent \textbf{#1. }}}

\definecolor{Gray}{gray}{0.9}
\newcommand{\modelName}{J-Moshi\xspace}

\title{Towards a Japanese Full-duplex Spoken Dialogue System}

\author{Atsumoto}{Ohashi}
\author{Shinya}{Iizuka}
\author{Jingjing}{Jiang}
\author{Ryuichiro}{Higashinaka}

\affiliation[nocounter]{Graduate School of Informatics}{Nagoya University}{Japan}

\email{
ohashi.atsumoto.c0@s.mail.nagoya-u.ac.jp
}

\usepackage{comment}

\begin{document}

\maketitle

\begin{abstract}
Full-duplex spoken dialogue systems, which can model simultaneous bidirectional features of human conversations such as speech overlaps and backchannels, have attracted significant attention recently. However, the study of full-duplex spoken dialogue systems for the Japanese language has been limited, and the research on their development in Japanese remains scarce. In this paper, we present the first publicly available full-duplex spoken dialogue model in Japanese, which is built upon Moshi, a full-duplex dialogue model in English. Our model is trained through a two-stage process: pre-training on a large-scale spoken dialogue data in Japanese, followed by fine-tuning on high-quality stereo spoken dialogue data. We further enhance the model's performance by incorporating synthetic dialogue data generated by a multi-stream text-to-speech system. Evaluation experiments demonstrate that the trained model outperforms Japanese baseline models in both naturalness and meaningfulness.\footnote{Our training codebase, fine-tuned models, and speech samples are available at \url{https://nu-dialogue.github.io/j-moshi}}
\end{abstract}

\section{Introduction}
Full-duplex spoken dialogue systems are attracting attention as a way to achieve natural voice interaction~\cite{nguyen-etal-2023-generative, wang2024a, ma2024language}. In dialogue, \emph{full-duplex} refers to having simultaneous bidirectional features such as speech overlaps and backchannels. Full-duplex spoken dialogue systems need to be researched for addressing the limitations of conventional \emph{half-duplex} dialogue systems~\cite{zhang-etal-2023-speechgpt, fang2025llamaomni} that wait for the other speaker to finish speaking before responding.

Moshi~\cite{defossez2024moshi} is a prominent full-duplex spoken dialogue model that achieves full-duplex dialogue by modeling both its own and the user's speech streams in parallel. Additionally, research on full-duplex spoken dialogue systems is increasing, particularly in English~\cite{veluri-etal-2024-beyond, zhang2024omniflatten, yu2024salmonn}. However, there is no full-duplex spoken dialogue system for other languages including Japanese, and insights into full-duplex spoken dialogue systems are lacking compared to in English.

The purpose of this study is to provide insights and baseline performance through the development of the first Japanese full-duplex spoken dialogue model. In this study, we adapt Moshi to Japanese through pre-training and fine-tuning using Japanese spoken dialogue data. In the pre-training stage, we use the J-CHAT corpus~\cite{nakata2024j}, which contains approximately 69,000 hours of monophonic spoken dialogue, to acquire basic capabilities of spoken dialogues in Japanese. Then, in the fine-tuning stage, we model full-duplex dialogue in Japanese using 344 hours of high-quality stereo spoken dialogue data where two speakers' voices are recorded on separate channels. Furthermore, we aim to improve dialogue capabilities through data augmentation using 602 hours of synthetic spoken dialogue generated by multi-stream text-to-speech (TTS). Through evaluation experiments, we demonstrate that the adapted model can acquire a certain level of Japanese speech capabilities. Moreover, through comparative analysis between English and Japanese, we show that the model can acquire Japanese-specific characteristics, such as having more speech overlaps~\cite{hayashi1988simultaneous, STUBBE1998257}.

\section{Moshi}
\label{sec:moshi}
\begin{figure}[t]
\centering
\includegraphics[scale=0.38]{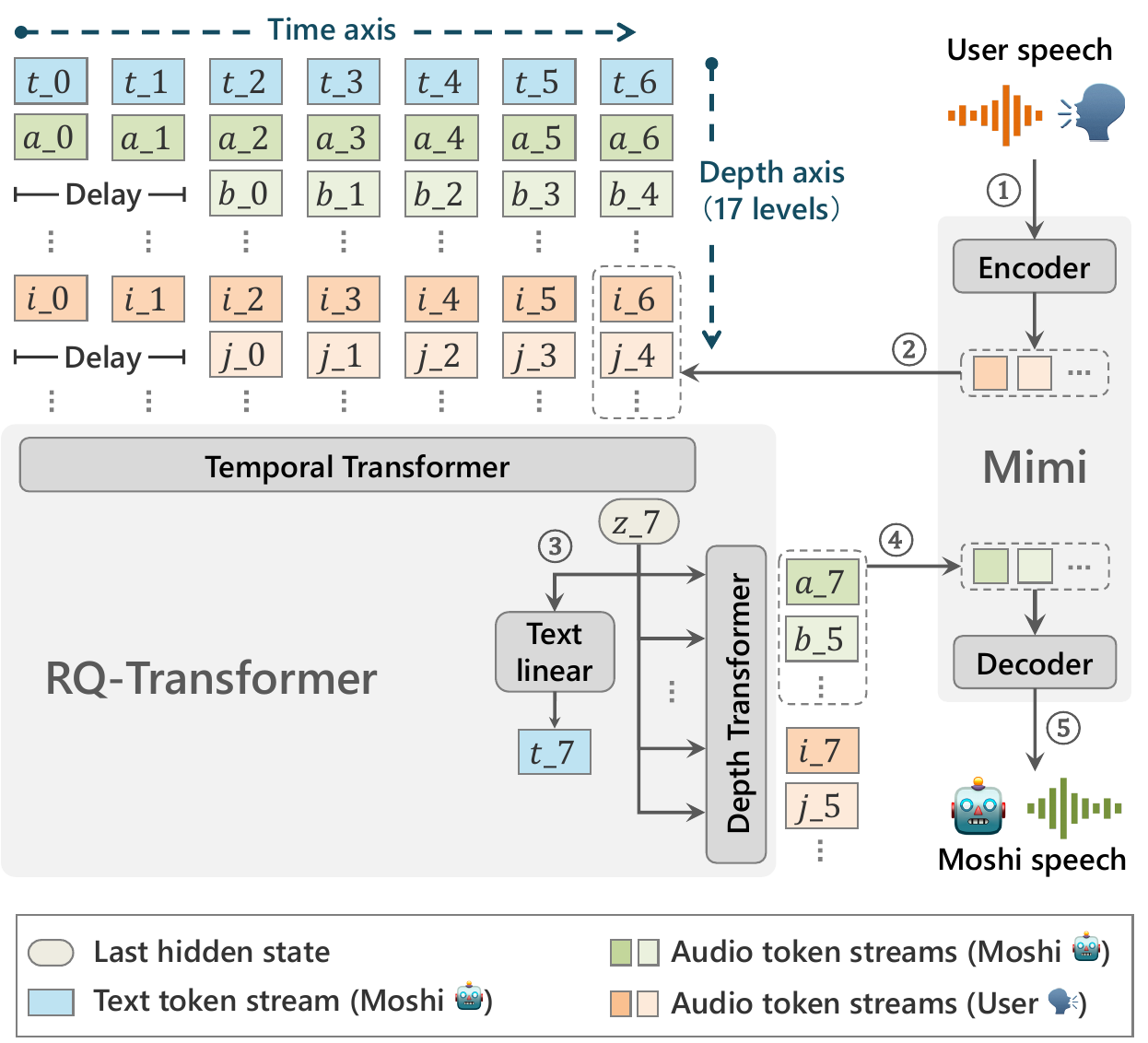}
\caption{Moshi's model architecture. It consists of the neural audio codec Mimi, which encodes speech waveforms into discrete audio tokens, and RQ-Transformer, which autoregressively models sequences of text tokens and audio tokens.}
\label{fig:moshi}
\vspace{-2mm}
\end{figure}

This section explains the model architecture of Moshi~\cite{defossez2024moshi} (Figure~\ref{fig:moshi}), which consists of the neural speech codec Mimi and the large-scale speech language model RQ-Transformer.

\subsection{Mimi}
Mimi is a neural speech codec~\cite{defossez2023high} consisting of a SEANet~\cite{tagliasacchi2020seanet} autoencoder and a residual vector quantizer~\cite{zeghidour2021soundstream}. The encoder discretizes 24,000Hz speech waveform data into audio tokens at a frame rate of 12.5Hz. To each frame, eight layers of tokens are assigned, where layer 1 is trained to retain semantic information of speech, while layers 2-8 are trained to retain acoustic information. A token of layer 1 is called the semantic token, and those of layers 2-8 are called acoustic tokens.

As shown in Figure~\ref{fig:moshi}, at each timestep (i.e., one frame), Mimi encodes the user's input speech to output eight user audio tokens and feeds them to RQ-Transformer. Simultaneously, it decodes eight Moshi audio tokens generated by RQ-Transformer to generate Moshi's output speech.

\subsection{RQ-Transformer}
RQ-Transformer consists of a Temporal Transformer based on a 7B-parameter large language model (LLM), which has been pre-trained with a large amount of web text, and a smaller Depth Transformer. This architecture allows Moshi to generate natural spoken dialogue by leveraging the high level of linguistic capabilities of the LLM.

The Temporal Transformer models token sequences along the time dimension at a rate of 12.5Hz. The token sequences include 17 layers: Moshi's text token sequence (1 layer), Moshi's audio token sequence (8 layers), and the user's audio token sequence (8 layers). Here, text tokens represent the textual content to be spoken as inner monologue. At each timestep $s$, it outputs a hidden vector $z_s$ from $(s-1)\times17$ tokens up to the previous step $s-1$. Then, the Text Linear layer samples the text token at $s$ from $z_s$. The Depth Transformer models audio tokens along the depth dimension at $s$. Taking $z_s$ as input, it autoregressively samples eight Moshi audio tokens and eight user audio tokens. Note that a delay of one timestep is applied to acoustic tokens to stabilize the quality of generated speech.

To align the timing of text tokens with audio tokens, token-level transcriptions created by Whisper~\cite{pmlr-v202-radford23a} (i.e., timestamp information of which timestep each text token corresponds to) are utilized, and PAD tokens are inserted into timesteps where no text tokens are assigned.

The training procedure of Moshi consists of pre-training with 7 million hours of monophonic audio data, fine-tuning with 2,000 hours of stereo spoken dialogue, and instruction tuning with 20,000 hours of stereo spoken dialogue synthesized by the multi-stream TTS system. For more detailed information about Moshi, refer to Moshi's technical paper~\cite{defossez2024moshi}.

\section{Training in Japanese}
\label{sec:training}

\begin{figure}[t]
\centering
\includegraphics[width=1\linewidth]{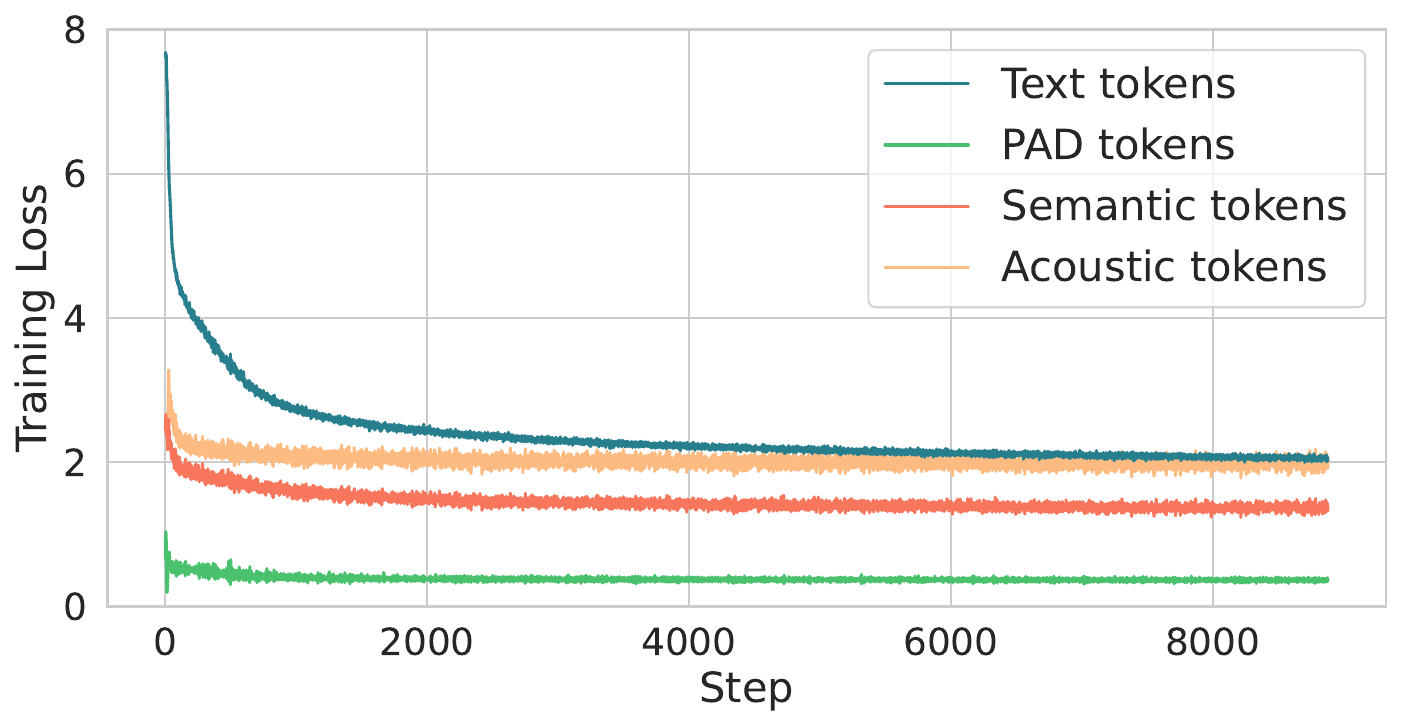}
\caption{Moshi's Loss curve during pre-training on J-CHAT}
\label{fig:training-curve}
\vspace{-2mm}
\end{figure}

In this study, we adapt Moshi to Japanese spoken dialogue. This section explains the text vocabulary adaptation and two-stage training steps, consisting of pre-training and fine-tuning, that were implemented to adapt original Moshi\footnote{\url{https://huggingface.co/kyutai/moshiko-pytorch-bf16}} to Japanese. We also explain data augmentation using multi-stream TTS. Through early experiments, we found that Mimi has a certain level of ability to encode and resynthesize Japanese speech without additional training, so we froze Mimi's parameters and trained only RQ-Transformer's parameters (Mimi's Japanese performance is shown in Section~\ref{sec:experiments}).

\subsection{Japanese Text Vocabulary Adaptation}
While Moshi's text tokenizer is a SentencePiece~\cite{kudo-richardson-2018-sentencepiece} trained on English data with a 32,000-word vocabulary, it does not include Japanese vocabulary and is therefore inefficient for tokenizing Japanese text. In this study, we adopted the SentencePiece model of Japanese GPT-2\footnote{\url{https://huggingface.co/rinna/japanese-gpt2-medium}} with a 32,000-word vocabulary as the Japanese tokenizer. Additionally, as we changed the vocabulary by replacing the tokenizer, we initialized some weights of RQ-Transformer that are tied to the text vocabulary. Specifically, we randomly initialized the text token embedding tables in the Temporal Transformer and Depth Transformer, as well as the Text Linear parameters.

\subsection{Pre-training}
\label{sec:pretraining}
The purpose of pre-training is to acquire fundamental capabilities of Japanese spoken dialogue through large-scale Japanese spoken dialogue data. In this study, we adopted the J-CHAT corpus~\cite{nakata2024j}, which contains 69,000 hours of Japanese spoken dialogue collected from YouTube and Podcasts.

\customParagraph{J-CHAT Preprocessing} Since J-CHAT's audio data is monophonic with all speakers' voices recorded in the same channel, it cannot be used directly for a two-channel spoken dialogue model (i.e., Moshi itself and the user). Therefore, following the method used for the original Moshi, we applied speaker diarization~\cite{Bredin23} to each audio file and created stereo spoken dialogues by randomly selecting one speaker for Moshi's channel and the others for the user's channel. Next, transcriptions for each channel were created using an automatic speech recognition (ASR) system.\footnote{\url{https://huggingface.co/reazon-research/reazonspeech-espnet-v2}} Furthermore, we obtained timestamps for each text token using WhisperX~\cite{bain2023whisperx} and inserted PAD tokens into timesteps without assigned text tokens in order to match the lengths of text token sequences and audio token sequences. The final data contained 3 billion text tokens, of which approximately 88\% were PAD tokens.

\customParagraph{Training on J-CHAT} We trained one epoch on the train set of the preprocessed J-CHAT (approximately 60,000 hours). We adopted ZeRO-3 parallelism implemented in the DeepSpeed library~\cite{10.5555/3433701.3433727} and conducted training on 128 NVIDIA V100 32GB GPUs. Mixed precision (float16) and activation checkpointing for each Transformer layer were used. The maximum length of each input sample was set to 2.7 minutes (2,048 tokens in the time direction), with a total batch size of 512 samples. Using AdamW~\cite{loshchilov2018decoupled} and following Llama 2 7B~\cite{touvron2023llama}, we set $\beta_1=0.9$, $\beta_2=0.95$, $\epsilon=1e-5$, and weight decay to $0.1$. The learning rate was set to $3e-5$ with 500 steps of linear warmup. Similar to the original Moshi, in loss calculation weighting, PAD token losses were reduced by 50\%, and the loss ratio between semantic tokens and acoustic tokens was set to $100:1$. The total number of optimization steps was $8,880$, requiring 36 hours. Figure~\ref{fig:training-curve} shows the loss curve during training.

\subsection{Fine-tuning}

\begin{table}[t]
\centering
\footnotesize
\caption{List of dialogue data used in this study}
\label{tab:corpus-list}
\begin{tabular}{llrr} \toprule
\multicolumn{2}{l}{Corpus} & \#Dialogues & Hours \\ \midrule
\rowcolor{Gray}\multicolumn{2}{l}{\textbf{Pre-training data}} & & \textbf{68,892} \\
 & J-CHAT~\cite{nakata2024j} & 4,937,497 & 68,892 \\ \midrule
\rowcolor{Gray}\multicolumn{2}{l}{\textbf{Fine-tuning data}} & & \textbf{344} \\
 & Japanese Callhome~\cite{LDC96S37} & 120 & 16 \\
 & CSJ~\cite{maekawa2003corpus} & 58 & 12 \\
 & Travel Agency Dialogue Corpus~\cite{inaba2024travel} & 330 & 115 \\
 & Casual Dialogue Corpus & 500 & 148 \\
 & Consultation Dialogue Corpus & 100 & 53 \\ \midrule
\rowcolor{Gray}\multicolumn{2}{l}{\textbf{Synthesized data by multi-stream TTS}} & & \textbf{602} \\
 & JapanesePersonaChat~\cite{10022973} & 4,983 & 94 \\
 & JapaneseEmpatheticDialogues~\cite{10022973} & 20,000 & 102 \\
 & Japanese Daily Dialogue~\cite{akama2023jdd-en} & 5,246 & 44 \\
 & RealPersonaChat~\cite{yamashita-etal-2023-realpersonachat} & 13,510 & 362 \\ \bottomrule
\end{tabular}
\vspace{-2mm}
\end{table}

\label{sec:finetuning}
The training data used in pre-training was forcibly divided into two channels from monophonic audio and therefore does not contain natural turn-taking such as speech overlaps and backchannels. To model actual full-duplex spoken dialogue, we performed fine-tuning using stereo spoken dialogue data where each speaker's voice was recorded in separate channels. We prepared a total of 344 hours of spoken dialogue data using five relatively large stereo spoken dialogue corpora in Japanese:
\begin{description}
\item[Japanese Callhome~\cite{LDC96S37}] A corpus of casual telephone conversations. We used the remaining 16 hours, excluding some audio with insufficient transcriptions.
\item[CSJ~\cite{maekawa2003corpus}] A corpus containing 660 hours of Japanese speech. We used only 12 hours of two-speaker spoken dialogue.
\item[Travel Agency Dialogue Corpus~\cite{inaba2024travel}] A corpus containing dialogues recorded via Zoom meetings between travel agency operator roles and customer roles. It contains a total of 115 hours of dialogue audio.
\item[Casual Dialogue Corpus (in-house)] A casual dialogue corpus created within our laboratory. Each dialogue was recorded via Zoom meetings. It contains a total of 148 hours of dialogue audio from 32 speakers.
\item[Consultation Dialogue Corpus (in-house)] A consultation dialogue corpus created within our laboratory, similar to the casual dialogue corpus. It contains a total of 53 hours of dialogue audio from 32 speakers.
\end{description}
Since the above corpora span a wide range of domains, we believe that the model can handle diverse domains. Table~\ref{tab:corpus-list} shows the breakdown of the data.

We tokenized the total 344 hours of the above data using the same method as in Section~\ref{sec:pretraining} and split it into train/valid/test sets at a ratio of 94:3:3. We then trained for 3 epochs on 320 hours of data included in the train set using 16 V100 32GB GPUs. The total number of optimization steps was $1,423$, requiring 2 hours. While basic hyperparameters were the same as during pre-training, the total batch size was set to 16 samples, and the learning rates for the Temporal Transformer and Depth Transformer were set to $2e-6$ and $4e-6$, respectively. Hereafter, we call the resulting model ``J-Moshi.'' 

\subsection{Data Augmentation using Multi-stream TTS}
\label{sec:multi-stream-tts}
The original Moshi was trained using 20,000 hours of stereo spoken dialogue synthesized from text dialogue using multi-stream TTS~\cite{defossez2024moshi}. We also synthesized spoken dialogue from text dialogue using multi-stream TTS and included it in the fine-tuning data. This is expected to increase the diversity of dialogues in the training data and acquire more versatile dialogue capabilities. Following D{\'e}fossez et al.'s implementation~\cite{defossez2024moshi}, we set \modelName's semantic token delay to 25 timesteps and acoustic token delay to 27 timesteps, then implemented the TTS model through pre-training on J-CHAT and fine-tuning on 344 hours of stereo spoken dialogue data. All training settings were the same as in Sections~\ref{sec:pretraining} and~\ref{sec:finetuning}. Below, we explain the text dialogue data and speech synthesis procedure.

\customParagraph{Text Dialogue Data Preparation} As text dialogue data for speech synthesis using multi-stream TTS, we used four existing text dialogue corpora: JapanesePersonaChat~\cite{10022973}, JapaneseEmpatheticDialogues~\cite{10022973}, Japanese Daily Dialogue Corpus~\cite{akama2023jdd-en}, and RealPersonaChat~\cite{yamashita-etal-2023-realpersonachat}. Since these corpora were collected in text chat, they contain many written language expressions and are not suitable for synthesizing spoken dialogue. Therefore, following previous research~\cite{fang2025llamaomni}, we used an LLM\footnote{\url{https://huggingface.co/google/gemma-2-27b-it}} to rewrite the text dialogues to include spoken language-specific expressions. As a result, we rewrote all 43,739 dialogues contained in the above four corpora.

\customParagraph{Stereo Spoken Dialogue Generation} Using multi-stream TTS, we generated 10 speech samples with different seed values for each text dialogue obtained in the previous step. Then, from these 10 samples, we selected the sample with the lowest Word Error Rate (WER) between the ASR result and the original dialogue text as the final speech sample for that dialogue. As a result, 602 hours of stereo spoken dialogue were synthesized. Table~\ref{tab:corpus-list} shows the breakdown of the synthesized speech data. The overall WER of the data was 24.6\%. 

We fine-tuned the model pre-trained on J-CHAT using a total of 946 hours of speech data, including 344 hours of the fine-tuning data and 602 hours of the synthesized speech data. The training settings were the same as in Section~\ref{sec:finetuning}, and the total number of optimization steps was $2,401$. Hereafter, we refer to this model trained on the augmented data as ``\modelName-ext.''

\section{Experiments}
\label{sec:experiments}
We adopted the prompted dialogue continuation task~\cite{nguyen-etal-2023-generative, veluri-etal-2024-beyond}, commonly used for evaluating full-duplex spoken dialogue models, to verify the spoken dialogue generation performance of \modelName and \modelName-ext through both automatic and human evaluations. This task involves generating a continuation from a few seconds of spoken dialogue prompt. We divided each dialogue in the test set of the fine-tuning data (see Section~\ref{sec:finetuning}) into 30-second chunks, and for each of the resulting 709 audio samples, we used the first 10 seconds as a prompt and had the model generate the subsequent 20 seconds.

\subsection{Baselines}
\label{sec:baselines}
As a comparison, we adopted dGSLM~\cite{nguyen-etal-2023-generative}, the typical full-duplex spoken dialogue model. For dGSLM training, like \modelName, we conducted pre-training using J-CHAT and fine-tuning using 344 hours of stereo spoken dialogue data. During inference, following Nguyen et al.'s settings~\cite{nguyen-etal-2023-generative}, we set top-$k$ to 20. Additionally, to evaluate Mimi's performance in Japanese spoken dialogue, we included simple re-synthesis of actual 20-second audio by Mimi (Re-synthesis). We also used the original 20-second audio (Ground-truth) as upper bound. As with the original Moshi, we tried three temperature parameters $\tau$ of 0.8, 0.9, and 1.0 during generation.

\subsection{Automatic Evaluation Results}
\label{sec:automatic-eval}
\begin{figure}[t]
\centering
\includegraphics[width=0.95\linewidth]{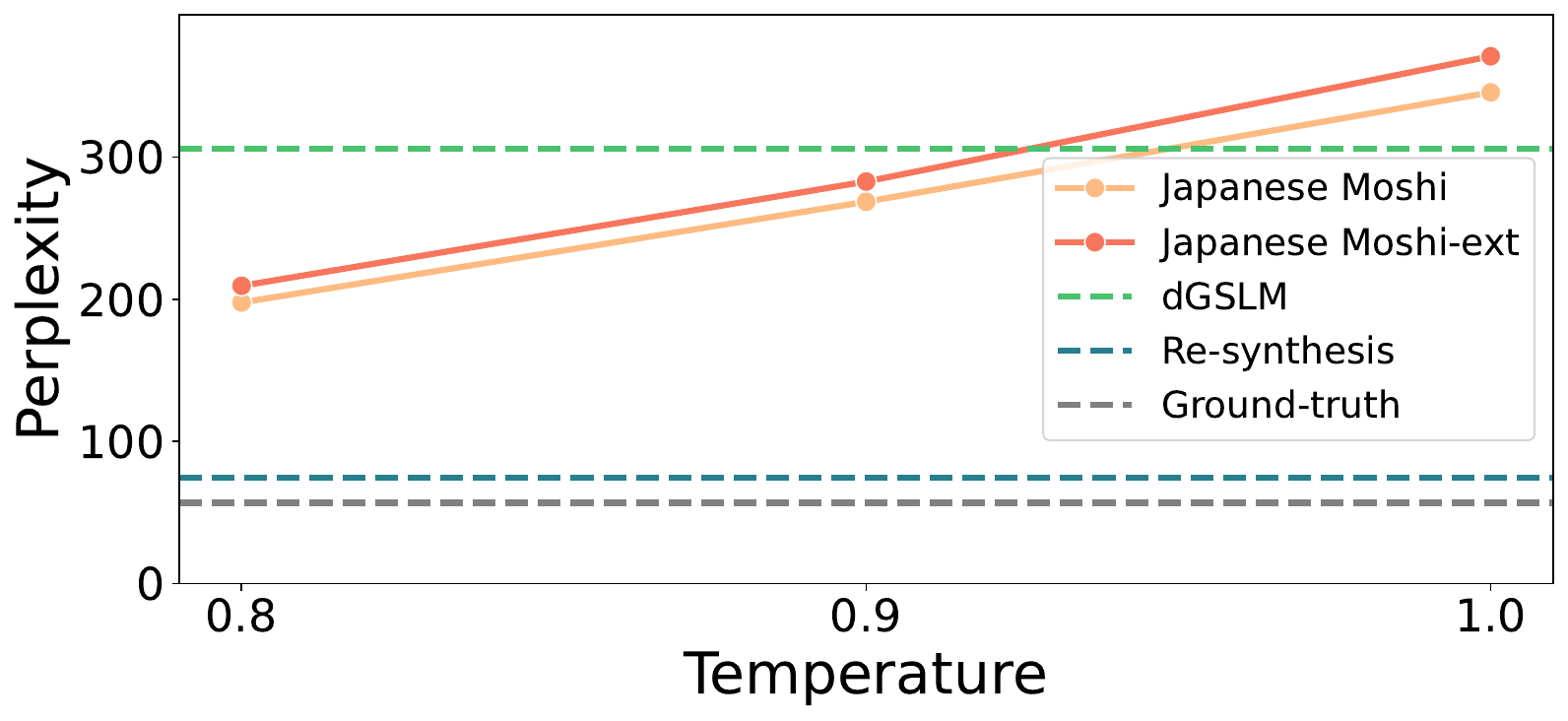}
\caption{Perplexity of speech generated by comparison models}
\label{fig:automatic-eval-result}
\vspace{-2mm}
\end{figure}

Following previous studies~\cite{nguyen-etal-2023-generative, defossez2024moshi}, we evaluated the model's fluency by measuring the perplexity (PPL) of a language model\footnote{\url{https://huggingface.co/llm-jp/llm-jp-3-3.7b}} on the ASR results of speech samples generated by each model. 

Results are shown in Figure~\ref{fig:automatic-eval-result}. The PPL of \modelName and \modelName-ext revealed that, similar to the original Moshi, lower $\tau$ values enable more fluent speech to be generated. Particularly with the $\tau=0.8$ setting, PPL improved by about 100 points compared to dGSLM trained from scratch on Japanese data. This suggests that adapting Moshi to Japanese positively affected speech fluency. However, compared to Re-synthesis and Ground-truth, the PPL significantly deteriorated, indicating the need for further improvements. Notably, Re-synthesis degraded minimally from Ground-truth. This suggests that Mimi can be applied directly to Japanese spoken dialogue for the first step of adapting Moshi to Japanese.

\subsection{Human Evaluation Results}
\label{sec:human-eval}
In the human evaluation\footnote{The experiment was approved by the ethical review committee of the Graduate School of Informatics, Nagoya University.}, we used 50 audio samples randomly selected from the 709 dialogue continuations of each model. For \modelName and \modelName-ext, we used $\tau=0.8$, which performed best in Section~\ref{sec:automatic-eval}. We recruited a total of 125 crowdworkers. Each worker evaluated ten samples. Following previous studies~\cite{nguyen-etal-2023-generative, nakata2024j}, they evaluated naturalness (how natural the dialogue sounds) and meaningfulness (how well the speech can be understood) on a 5-point scale.

Table~\ref{tab:human-eval-result} shows the results. For naturalness, both Japanese Moshi and \modelName-ext outperformed dGSLM. They also significantly surpassed dGSLM in meaningfulness. Notably, \modelName-ext further improved meaningfulness compared to \modelName, indicating that dialogue data augmentation through multi-stream TTS contributed to improving language capability. However, compared to Re-synthesis, both naturalness and meaningfulness deteriorated by more than 1 point, suggesting RQ-Transformer itself has significant room for improvement. Additionally, Re-synthesis scores degraded by about 0.5 points compared to Ground-truth, indicating that Mimi will also need to be adapted to Japanese in the future.

\begin{table}[t]
\centering
\small
\caption{Five-point evaluation scores and 95\% confidence intervals for generated spoken dialogues. $\tau$ indicates the temperature parameter during generation.}
\label{tab:human-eval-result}
{\tabcolsep=1.5mm
\begin{tabular}{lccc} \toprule
Model & $\tau$ & Naturalness & Meaningfulness \\ \midrule
dGSLM & & 2.44$\pm$0.12 & 1.76$\pm$0.09 \\
\modelName & 0.8 & \textbf{2.67$\pm$0.13} & 2.19$\pm$0.12 \\
\modelName-ext & 0.8 & 2.66$\pm$0.13 & \textbf{2.30$\pm$0.13} \\ \midrule
Re-synthesis & & 3.90$\pm$0.12 & 3.92$\pm$0.13 \\
Ground-truth & & 4.46$\pm$0.09 & 4.45$\pm$0.10 \\ \bottomrule
\end{tabular}
}
\end{table}

\section{Comparison of Japanese and English}
To analyze what kind of Japanese-specific behaviors \modelName acquired through our training, we compared it with the original Moshi. Specifically, we used four turn-taking statistics~\cite{nguyen-etal-2023-generative}: Inter-Pausal Units (IPU) (speech segments separated by at least 0.2 seconds of silence), Pause (silence between IPUs from the same speaker), Gap (silence between IPUs from different speakers), and Overlap (time where IPUs from different speakers overlap). For the 709 dialogue continuations generated by \modelName in Section~\ref{sec:experiments}, we calculated each measure as an average per minute. For the original Moshi, we used the values reported by D{\'e}fossez et al.~\cite{defossez2024moshi}. While the experimental conditions are not strictly identical, the comparison is considered valid as key settings such as temperature parameters are the same.

\begin{table}[t]
\centering
\footnotesize
\caption{Turn-taking metrics per minute in generated spoken dialogues. $\tau$ indicates the temperature parameter during generation. $\dagger$ indicates values reported by D{\'e}fossez et al.~\cite{defossez2024moshi}.}
\label{tab:analysis-result}
{\tabcolsep=0.8mm
\begin{tabular}{llcccccc} \toprule
Language & Model & Samples & $\tau$ & IPU & Pause & Gap & Overlap \\ \midrule
\multirow{3}{*}{Japanese} & J-Moshi & 709 & 0.8 & 53.2s & 6.3s & 4.5s & \textbf{5.0s} \\
 & J-Moshi-ext & 709 & 0.8 & 50.9s & 7.0s & 4.6s & 4.2s \\
 & Ground-truth & 709 & & 59.7s & 3.5s & 4.0s & 8.1s \\ \midrule
\multirow{2}{*}{English} & Moshi$^\dagger$ & 1,000 & 0.8 & 35.1s & 13.2s & 12.5s & 1.2s \\ 
 & Ground-truth$^\dagger$ & 1,000 & & 51.1s & 6.4s & 4.2s & 3.3s \\ \bottomrule
\end{tabular}
}
\vspace{-2mm}
\end{table}

Results are shown in Table~\ref{tab:analysis-result}. We can see that spoken dialogues generated by \modelName have more IPUs and overlaps than Moshi. It has been reported that Japanese dialogues have more overlaps and backchannels than English~\cite{hayashi1988simultaneous, STUBBE1998257}, and this tendency is supported by Ground-truth shown in Table~\ref{tab:analysis-result}. These results suggest that \modelName has acquired Japanese-specific turn-taking characteristics. Another difference between \modelName and Moshi is the sparseness of text tokens relative to audio tokens. While it was reported that the PAD token ratio was 65\% in Moshi's training data, the PAD token ratio in J-CHAT was 88\% as we found in Section~\ref{sec:pretraining}. This is likely due to the characteristics of Japanese where each token contains more phonemes, such as kanji characters, making the density of text tokens sparse relative to audio. This suggests the need for learning settings and objective functions that consider Japanese characteristics, such as adjusting the loss weights for PAD tokens. 

\section{Conclusion}
In this study, we developed \modelName, a Japanese adaptation of Moshi, as a Japanese full-duplex spoken dialogue model. We conducted pre-training with 60,000 hours of Japanese speech data and fine-tuning with stereo spoken dialogue data and further attempted to improve performance using synthetic data generated through multi-stream TTS. In our experiments, we evaluated the naturalness and meaningfulness of dialogue speech generated by \modelName. Furthermore, through comparative analysis with Moshi, we demonstrated that \modelName acquired Japanese-specific behavior such as increased speech overlaps. We hope our findings will contribute to research and development of full-duplex spoken dialogue systems across multiple languages.

\section{Acknowledgements}
This work was supported by JST Moonshot R\&D, Grant number JPMJMS2011. We used the computational resources of the supercomputer ``Flow'' at the Information Technology Center, Nagoya University.

\bibliographystyle{IEEEtran}
\bibliography{mybib}

\end{document}